\documentclass{article}

\usepackage{pgfplots}
\pgfplotsset{compat=1.17}
\usepackage{microtype}
\usepackage{graphicx}
\usepackage{subfigure}
\usepackage{booktabs} % for professional tables

\usepackage{hyperref}

\usepackage{amsmath,amsthm}
\usepackage{amssymb}
\usepackage{multirow}
\usepackage{nicefrac}
\usepackage{pifont}
\usepackage[accepted]{icml2021}

\icmltitlerunning{Weight-Covariance Alignment for Adversarially Robust Neural Networks}

\begin{document}
\twocolumn[
\icmltitle{Weight-Covariance Alignment for Adversarially Robust Neural Networks}




\begin{icmlauthorlist}
\icmlauthor{Panagiotis Eustratiadis}{uoe}
\icmlauthor{Henry Gouk}{uoe}
\icmlauthor{Da Li}{uoe,samsung}
\icmlauthor{Timothy Hospedales}{uoe,samsung}
\end{icmlauthorlist}

\icmlaffiliation{uoe}{University of Edinburgh}
\icmlaffiliation{samsung}{Samsung AI Center, Cambridge}

\icmlcorrespondingauthor{Panagiotis Eustratiadis}{p.eustratiadis@ed.ac.uk}
\icmlcorrespondingauthor{Henry Gouk}{henry.gouk@ed.ac.uk}
\icmlcorrespondingauthor{Da Li}{dali.academic@gmail.com}
\icmlcorrespondingauthor{Timothy Hospedales}{t.hospedales@ed.ac.uk}

\icmlkeywords{Stochastic Neural Network, Adversarial Robustness, Machine Learning}

\vskip 0.3in
]


\printAffiliationsAndNotice{}  % leave blank if no need to mention equal contribution

\newcommand{\alg}{WCA-Net}
\newcommand{\xmark}{\ding{55}}%
\newcommand{\myparagraph}[1]{\noindent\textbf{#1}}\quad 
\newtheorem{theorem}{Theorem}

\begin{abstract}
Stochastic Neural Networks (SNNs) that inject noise into their hidden layers have recently been shown to achieve strong robustness against adversarial attacks. However, existing SNNs are usually heuristically motivated, and often rely on adversarial training, which is computationally costly. We propose a new SNN that achieves state-of-the-art performance without relying on adversarial training, and enjoys solid theoretical justification. Specifically, while existing SNNs inject learned or hand-tuned isotropic noise, our SNN learns an anisotropic noise distribution to optimize a learning-theoretic bound on adversarial robustness. We evaluate our method on a number of popular benchmarks, show that it can be applied to different architectures, and that it provides robustness to a variety of white-box and black-box attacks, while being simple and fast to train compared to existing alternatives.
\end{abstract}

\section{Introduction}
\label{sec:introduction}

It has been shown that deep convolutional neural networks, while displaying exceptional performance in computer vision problems such as image recognition~\cite{cvpr16resnet}, are vulnerable to input perturbations that are imperceptible to the human eye~\cite{iclr14intriguing}. The perturbed input images, known as adversarial examples, can be generated by single-step~\cite{iclr15fgsm} and multi-step~\cite{iclr18pgd,iclr2016bim, sp17cnw} updates using both gradient-based optimization methods and derivative-free approaches~\cite{acmais2017zoo}. This vulnerability raises the question of how one can go about ensuring the security of machine learning systems, thus preventing a malicious entity from exploiting instabilities~\cite{tkde2013security}. In order to tackle this problem, many adversarial defense algorithms have been proposed in the literature. Among them, Stochastic Neural Networks (SNNs) that inject fixed or learnable noise into their hidden layers have shown promising results~\cite{eccv18rse,iclr19advbnn,iccv19pni,cvpr20learn2perturb,aaai2021sesnn}.

In this paper, we identify three limitations of the current state-of-the-art stochastic defense methods. First, most contemporary adversarial defense methods use a mixture of clean and adversarial (or even purely adversarial) samples during training, i.e., adversarial training~\cite{iclr15fgsm, iclr18pgd, iclr19advbnn, iccv19pcl, iccv19pni, cvpr20learn2perturb}. However, generating strong adversarial examples during training leads to significantly higher computational cost and longer training time.
% More importantly, adversarial training assumes that the attack method is already known. However, it has been demonstrated that, though effective in improving robustness against a specific attack, maintaining robustness across different \emph{unseen} attacks is a challenge~\cite{nips19free}.
Second, many existing adversarial defenses~\cite{iccv19pcl}, and especially stochastic defenses~\cite{cvpr20learn2perturb} are heuristically motivated. Although they may be empirically effective against existing attacks, they lack theoretical support. Third, the noise incorporated by existing stochastic models is \emph{isotropic} (i.e., generated from a multivariate Gaussian distribution with a diagonal covariance matrix), meaning that it perturbs the learned features of different dimensions independently. Our theoretical analysis will show that this is a strong assumption and best performance is expected from \emph{anisotropic} noise.

We address the aforementioned limitations and propose an SNN that makes use of learnable anisotropic noise. We theoretically analyse the margin between the clean and adversarial performance of a stochastic model and derive an upper bound on the difference between these two quantities. This novel theoretical insight suggests that the anisotropic noise covariance in an SNN should be optimized to align with the classifier weights, which has the effect of tightening the bound between clean and adversarial performance. This leads to an easy-to-implement regularizer, which can be efficiently optimized on clean samples alone without need for adversarial training. We show that our method, called Weight-Covariance Alignment (WCA), can be applied to architectures of varied depth and complexity (namely, LeNet++ and ResNet-18), and achieves state-of-the-art robustness across several widely used benchmarks, including CIFAR-10, CIFAR-100, SVHN and F-MNIST. Moreover, this high level of robustness is demonstrated for both white-box and black-box attacks. We name our proposed model \alg{}.

The contributions of our paper are summarized as follows:
\begin{itemize}
  \item While the majority of existing stochastic defenses are heuristically motivated, our proposed method is derived by optimizing a learning theoretic bound, providing solid justification for its robust performance.
  \item To the best of our knowledge, we are the first to propose a stochastic defense with learned anisotropic noise.
  \item WCA only requires clean samples for training, unlike most of the current state-of-the art defenses that depend on costly adversarial training. %Additionally, it does not introduce any new hyperparameters that require tuning.
  \item We demonstrate the state-of-the-art performance of our method on various benchmarks and provides resilience to both white- and black-box attacks.
\end{itemize}

\section{Related Work}
\label{sec:related-work}

\subsection{Adversarial Attacks}
We consider the standard threat model, where the attacker can construct norm-bounded perturbations to a clean input. %First-order adversaries, also known as white-box attacks, perturb the input image of a target model towards the direction that deceives the target model to misclassify the input.
First-order white-box adversaries use the gradient with respect to the input image to perturb it in the direction that increases misclassification probability. 
The attack can also be targeted or untargeted, depending on whether a specific  misclassification is required~\cite{iclr15fgsm, iclr2016bim, iclr18pgd, sp17cnw}. By default, we consider the untargeted variants of these attacks. The simplest first-order adversary is the Fast Gradient Sign Method (FGSM), proposed in~\citet{iclr15fgsm}. The attack adds a small perturbation to the input in the direction indicated by the sign of the gradient of the classification loss, $\mathcal{L}$, w.r.t. the input, $\vec x$, controlled by an attack strength $\epsilon$,
\begin{equation*}
    \vec x^\prime = \vec x + \epsilon \cdot \operatorname{sign}(\nabla_{\vec x} \mathcal{L}(h(\vec x), y)),
\end{equation*}
where $h$ is the target model. \citet{iclr2016bim} upgraded this single-step attack to a multi-step version named Basic Iterative Method (BIM) with iterative updates and smaller step size at each update. Though BIM works effectively, \citet{iclr18pgd} demonstrated that randomly initializing the perturbation generated by BIM, and then making multiple attempts to construct an adversarial example results in a stronger adversarial attack known as Projected Gradient Descent (PGD). Another white-box attack of slightly different nature is the C\&W attack~\cite{sp17cnw}, which aims to find an input perturbation $\delta$ that maximizes the following objective:

\begin{equation*}
\begin{aligned}
    \mathcal{L}(h(\vec x + \delta), y) - ||\delta||_{p}\\
    s.t. \quad \vec x + \delta\in [0, 1]^{n},
\end{aligned}
\end{equation*}
where $p$ is commonly chosen from $\{0,2,\infty\}$.

Different from the white-box attacks, black-box attacks assume the details of the targeted model are unknown, and one can only access the model through queries. Therefore, in order to attack a target model in this case, one typically trains a substitute of it~\cite{acm2017practicalbbox} and generates an attack using the queried prediction of the target model and the local substitute. Also, instead of training a substitute for the target model, zero-order optimization methods~\cite{acmais2017zoo,ec19onepixel} have been proposed to estimate the gradients of the target model directly.
In this paper, we demonstrate that our proposed method is robust against both white- and black-box attacks.

\subsection{Stochastic Adversarial Defense}

Recent work has shown that SNNs yield promising performance in adversarial robustness. This can be achieved by injecting either fixed~\cite{eccv18rse} or learnable~\cite{iccv19pni, cvpr20learn2perturb, aaai2021sesnn} noise into the models.

The idea behind Random Self Ensemble (RSE)~\cite{eccv18rse} is that one can simulate an ensemble of virtually infinite models while only training one. This is achieved by injecting additive spherical Gaussian noise into various layers of a network and performing multiple forward passes at test time. Though simple, it effectively improves the model robustness in comparison to a conventional deterministic model.
RSE treats the variance of the injected noise as a hyperparameter that is heuristically tuned, rather than learned in conjunction with the other network parameters. In contrast, \citet{iccv19pni} propose Parametric Noise Injection (PNI), where a fixed spherical noise distribution is controlled by a learnable ``intensity'' parameter, further improving model robustness. The authors show that the noise can be incorporated into different locations of a neural network, i.e., it is applicable to both feature activations and model weights. The injected noise is trained together with the model parameters via adversarial training.
Learn2Perturb (L2P)~\cite{cvpr20learn2perturb} is a recent extension of PNI. Instead of learning a single spherical noise parameter, L2P learns a set of parameters defining an isotropic noise perturbation-injection module. The parameters of the perturbation-injection  module and the model are updated alternatingly in a manner named ``alternating back-propagation'' by the authors, using adversarial training.
Finally, SE-SNN~\cite{aaai2021sesnn} introduces fully-trainable stochastic layers, which are trained for adversarial robustness by adding a regularization term to the objective function that maximizes the entropy of the learned noise distribution. Unlike the other SNNs, but similarly to ours, SE-SNN only requires clean training samples. 

Although conceptually related to the aforementioned stochastic defense methods, \alg{} differs in several important aspects: \alg{} is the first stochastic model to inject learnable \emph{anisotropic} noise into the latent features. Our approach is derived from from optimization of a learning theoretic bound on the adversarial generalisation performance of SNNs, which motivates the use of anisotropic noise. \alg{} does not require adversarial training and can be optimized with clean samples alone, and is therefore simpler and more efficient to train.

Another class of stochastic defenses apply noise to the input images, rather than injecting noise to intermediate activations~\citep{pinot2019, cohen2019, li2019, lee2019}. From a theoretical point of view, this can be seen as ``smoothing'' the function implemented by the neural network in order to reduce the amount the output of the network can change when the input is changed only slightly. This type of defense can be considered a black-box defense, in the sense that it does not actually involve regularizing the weights of the network --- it only modifies the input. While interesting, it has primarily been applied in scenarios where one is using a model-as-a-service framework, and cannot be sure if the model was trained with some sort of adversarial defense or not~\citep{cohen2019}.

\section{Methods}
\label{sec:methods}

Based on theoretical analysis of how the injected noise can impact generalisation performance, further expanded in Section~\ref{sec:wca}, we propose a weight-covariance alignment loss term that encourages the weight vectors associated with the final linear classification layer to be aligned with the covariance matrix of the injected noise. Consequently, our theory leads us to use anisotropic noise, rather than the isotropic noise typically employed by previous approaches.

Our method fits into the family of SNNs that apply additive noise to the penultimate activations of the network. Consider the function, $f(\vec x)$, which implements the feature extractor portion of the network i.e., everything except the final classification layer. Our \alg{} architecture is defined as
\begin{equation*}
    h(\vec x) = W (f(\vec x) + \vec z) + \vec b, \;\; \vec z \sim \mathcal{N}(0, \Sigma),
\end{equation*}
where $W$ and $\vec b$ are the parameters of the final linear layer, $\vec z$ is the vector of additive noise. The objective function used to train this model is
\begin{equation}
    \label{eq:loss_func}
    \mathcal{L} = \mathcal{L}_{\text{C}} - \mathcal{L}_{\text{WCA}},
\end{equation}
where $\mathcal{L}_{\text{C}}$ and $\mathcal{L}_{\text{WCA}}$ represent the classification loss (e.g. softmax composed with cross entropy) and weight-covariance alignment term respectively. We describe each of our technical contributions in the remainder of this Section.

\subsection{Weight-Covariance Alignment}
\label{sec:wca}
Non-stochastic methods for defending against adversarial examples typically try to guarantee that the prediction for an input image cannot be changed. In contrast, a defense that is stochastic should aim to minimize the probability that the prediction can be changed. In this Section, we present a theoretical analysis of the probability that the prediction of an SNN will be changed by an adversarial attack. For simplicity, we restrict our analysis to the case of binary classification.

Denoting a feature extractor as $f$, we define an SNN $h$, trained for binary classification as
\begin{equation*}
    h(\vec x) = \vec w^T (f(\vec x) + \vec z) + b, \;\; \vec z \sim \mathcal{N}(0, \Sigma),
\end{equation*}
where $\vec w$ is the weight vector of the classification layer and $b$ is the bias. We denote the non-stochastic version of $h$, where the value of $\vec z$ is always a vector of zeros, as $\Tilde{h}$. The margin of a prediction is given by
\begin{equation*}
    m_h(\vec x, y) = y h(\vec x),
\end{equation*}
for $y \in \{-1, 1\}$. It is positive if the prediction is correct and negative otherwise.

The quantity in which we are interested is the difference in probabilities of misclassification when the model is and is not under adversarial attack $\delta$, which is given by
\begin{equation}
    \label{eq:theory_adv_gap}
    \begin{split}
    G_{p,\epsilon}^h(\vec x, y) = \max_{\vec \delta : \|\vec \delta\|_p \leq \epsilon} P(m_h(\vec x + \delta, y) \leq 0)\\ - P(m_h(\vec x, y) \leq 0).
    \end{split}
\end{equation}
Our main theoretical result, given below, shows how one can take an adversarial robustness bound, $\Delta_p^{\Tilde{h}}(\vec x, \epsilon)$, for the deterministic version of a network, and transform it to a bound on $G$ for the stochastic version of the network.

\begin{theorem}
\label{thm:bound}
The quantity $G_{p,\epsilon}^h(\vec x, y)$, as defined above, is bounded as
\begin{equation*}
    G_{p,\epsilon}^h(\vec x, y) \leq \frac{\Delta_p^{\Tilde{h}}(\vec x, \epsilon)}{\sqrt{2 \pi\vec w^T \Sigma \vec w }},
\end{equation*}
where the robustness of the deterministic version of $h$ is known to be bounded as $|\Tilde{h}(\vec x) - \Tilde{h}(\vec x + \vec \delta)| \leq \Delta_p^{\Tilde{h}}(\vec x, \epsilon)$ for any $\|\vec \delta\|_p \leq \epsilon$.
\end{theorem}
The proof is provided in the supplemental material. We can see from Theorem~\ref{thm:bound} that increasing the bi-linear form, $\vec w^T \Sigma \vec w$, of the noise distribution covariance and the classifier reduces the gap between clean and robust performance. As such, we define the loss term,
\begin{equation}
    \label{eq:loss_wca}
    \mathcal{L}_{\text{WCA}} = \sum_{i=1}^C \text{ln}(\vec w_i^T \Sigma \vec w_i),
\end{equation}
where $C$ is the number of classes in the classification problem, and $\vec w_i$ is the weight vector of the final layer that is associated with class $i$. We found that including the logarithm results in balanced growth rates between the $\mathcal{L}_C$ and $\mathcal{L}_{\text{WCA}}$ terms in Eq.~\ref{eq:loss_func} as training progresses, hence improving the reliability of training loss convergence.

The key insight of Theorem~\ref{thm:bound}, operationalized by Eq.~\ref{eq:loss_wca}, is that the noise and weights should co-adapt to align the noise and weight directions. We call this loss Weight-Covariance Alignment (WCA) because it is maximized when each $\vec w_i$ is well-aligned with the eigenvectors of the covariance matrix.

This WCA loss term runs into the risk of maximizing the magnitude of $\vec w$, rather than encouraging alignment or increasing the scale of the noise. To avoid the uncontrollable scaling of network parameters, it is common practice to penalize large weights by means of $\ell^2$ regularization:
\begin{equation*}
    \mathcal{L} = \mathcal{L_C} - 
    \mathcal{L}_{WCA}+\lambda \vec w^T \vec w,
\end{equation*}
where $\lambda$ controls the strength of the penalty. In our case, we apply the $\ell^2$ penalty when updating the parameters of the classification layer and the covariance matrix. Another approach to limiting parameter magnitude would be to enforce norm constraints on $\vec{w}$ and $\Sigma$, e.g., using a projected subgradient method at each update. We provide more details of this alternative in the supplementary material. Empirically, we found that the penalty-based approach outperformed the constraint-based approach, so we focus on the former by default. 

\subsection{Injecting Anisotropic Noise}
In contrast to previous work that only considers injecting isotropic Gaussian noise  \cite{iclr19advbnn,iccv19pni,cvpr20learn2perturb,aaai2021sesnn}, we make use of anisotropic noise, providing a richer noise distribution than previous approaches. Crucially, it also means that the principal directions in which the noise is generated no longer have to be axis-aligned. I.e., prior work suffers from the inability to simultaneously optimise alignment between noise and weights (required to minimise the adversarial gap bounded by Theorem~\ref{thm:bound}), and freedom to place weight vectors off the axis (required for good clean performance). Our use of anisotropic noise in combination with WCA encourages alignment of the weight vectors with the covariance matrix eigenvectors, while allowing non-axis aligned weights, thus providing more freedom about where to place the classification decision boundaries.

Previous approaches are able to train the variance of each dimension of the isotropic noise via the use of the ``reparameterization trick'' \cite{kingma2014vae}, where one samples noise from a distribution with zero mean and unit variance, then rescales the samples to get the desired variance. Because the rescaling process is differentiable, this allows one to learn variance jointly with the other network parameters with backpropagation. In order to sample anisotropic noise, one can instead sample a vector of zero mean unit variance and multiply this vector by a lower triangular matrix, $L$. This lower triangular matrix is related to the covariance matrix as
\begin{equation*}
    \Sigma = L \cdot L^T.
\end{equation*}
This guarantees that the covariance matrix remains positive semi-definite after each gradient update.

\section{Experiments}
\label{sec:experiments}
In this Section we present the experiments that demonstrate the efficacy of our model and verify our theoretical analysis.
\subsection{Experimental Setup}

%\subsubsection{Datasets}

\myparagraph{Datasets} For comparison against the current state-of-the-art and for our ablation study we use four benchmarks: CIFAR-10, CIFAR-100~\cite{krizhevsky2009learning}, SVHN~\cite{netzer2011reading} and Fashion-MNIST~\cite{corr17fmnist}. CIFAR-10 and CIFAR-100 contain 60K 32x32 color images, 50K for training and 10K for testing, evenly spread across 10 and 100 classes respectively. SVHN can be considered a more challenging version of MNIST~\cite{lecun2010mnist}; it contains almost 100K 32x32 color images of digits (0-9) collected from Google's Street View imagery, with roughly 73K for training and 26K for testing. Fashion-MNIST is a collection of 70K 28x28 grayscale images of clothing, 60K for training and 10K for testing, also spread across 10 classes.

\myparagraph{Models} For all benchmarks except F-MNIST we use a ResNet-18~\cite{cvpr16resnet} backbone, while for F-MNIST, being a relatively simpler dataset, we use LeNet++~\cite{eccv16lenet}. After the backbone we add a penultimate layer for dimensionality reduction; this enables us to always train a reasonably-sized covariance matrix regardless of the original dimensionality of the feature extractor\footnote{32x32 for the benchmarks with 10 classes, 256x256 for the benchmarks with 100 classes.}. The only restriction for the dimensionality of the penultimate layer is that it needs to be a number greater or equal to the number of classes in the task, so as to allow the covariance matrix to align with at least one classifier vector. The two hyperparameters of note across all of our experiments are the learning rate and $\ell^2$ penalty (i.e., weight decay), the exact values of which are provided in the supplementary material. 

\subsubsection{Attacks}

We evaluate our method using three white-box adversaries: FGSM~\cite{iclr15fgsm}, PGD~\cite{iclr18pgd} and C\&W~\cite{sp17cnw}, and one black-box attack: the One-Pixel attack~\cite{ec19onepixel}. 

We parameterize the attacks following the literature~\cite{iccv19pni, cvpr20learn2perturb}. More specifically, FGSM and PGD are set with an attack strength of $\epsilon=8/255$ for CIFAR-10, CIFAR-100 and SVHN, and $\epsilon=0.3$ for F-MNIST. PGD has a step size of $\alpha=\epsilon/10$ and number of steps $k=10$ for all benchmarks as per~\citet{iccv19pni}. C\&W has a learning rate of $\alpha=5\cdot10^{-4}$, number of iterations $k=1000$, initial constant $c=10^{-3}$ and maximum binary steps $b_{\text{max}} = 9$ same as~\citet{cvpr20learn2perturb}.

For the parameters of the One-Pixel attack we tried to replicate the experimental setup described in the supplementary material of~\citet{cvpr20learn2perturb} for attack strengths of 1, 2 and 3 pixels. We followed their setup with population size $\mathrm{N=400}$ and maximum number of iterations $\mathrm{k_{max}=75}$. However, we noticed that the more pixels we added to our attack the weaker the attack became, which is counter-intuitive. We attribute that to the small number of iterations; every added pixel substantially increases the search space of the differential evolution algorithm, and 75 iterations are no longer enough to converge when the number of pixels is 2 and 3. Therefore we maintain a population size of $\mathrm{N=400}$, but increase the number of iterations to $\mathrm{k_{max}=1000}$. For reproducibility purposes, we further clarify that for the differential evolution algorithm we use a crossover probability of $\mathrm{r=0.7}$, a mutation constant of $\mathrm{m=0.5}$, and the following criterion for convergence:
\begin{equation*}
    \sqrt{\text{Var}(\mathcal{E}(X))} \leq \Big | \frac{1}{100N}\sum_{x \in X} \mathcal{E}(x) \Big |,
\end{equation*}
where $X$ denotes the population, $\mathcal{E}(X)$ the energy of the population and $\mathcal{E}(x)$ the energy of a single sample.

\myparagraph{Expectation over Transformation}
Due to the noise injected by SNNs, the gradients used by white-box attacks are stochastic  \cite{pmlr18obfuscated}. 
As a result, the true gradients cannot be correctly estimated for attacks that use only one sample to compute the perturbation. To avoid this issue, we apply Expectation over Transformation (EoT) following~\citet{pmlr18obfuscated}. When generating an attack, we compute gradients of multiple forward passes using Monte-Carlo sampling and perturb the inputs using the averaged gradient at each update. We empirically found that a reliable number of MC samples is 50 (as we observed performance begins to saturate from around 35 and converges at 40); thus, we use 50 across all experiments.

\subsection{Comparison to Prior Stochastic Defenses}
\myparagraph{Competitors}
We compare the performance of \alg{} to three recent state-of-the-art stochastic defenses to verify its efficacy. \textbf{AdvBNN}~\cite{iclr19advbnn}: adversarially trains a Bayesian neural network for defense. \textbf{PNI}~\cite{iccv19pni}: learns an ``intensity'' parameter to control the variance of their SNN. \textbf{Learn2Perturb (L2P)}~\cite{cvpr20learn2perturb}: improves PNI by learning an isotropic perturbation injection module. Furthermore, there are partial comparisons against \textbf{SE-SNN}~\cite{aaai2021sesnn} and \textbf{IAAT}~\cite{cvpr19denoising}. All experiments use a ResNet-18 backbone and are conducted on CIFAR-10 for fair comparison.

% \begin{table}[t]
%     \caption{Comparison of state-of-the-art SNNs for FGSM and PGD attacks on CIFAR-10 with a ResNet-18 backbone. Performance of competing methods extracted from~\citet{cvpr20learn2perturb}.}
%     \centering
%     \vskip 0.15in
%     \begin{tabular}{lccc}
%         \toprule
%         Method      & Clean & FGSM  & PGD \\
%         \midrule
%         Adv-BNN     & 82.15 & 60.04 & 53.62 \\
%         PNI         & \textbf{87.21} & 58.06 & 49.42 \\
%         L2P         & 85.30 & 62.43 & 56.06 \\
%         \alg{}      & 85.40 & \textbf{72.47} & \textbf{71.71} \\
%         \bottomrule
%     \end{tabular}
%     \label{tab:sota_compare_fgsm_pgd_cifar10}
% \end{table}

% \begin{table}[t]
%     \caption{Comparison of state-of-the-art SNNs for FGSM and PGD attacks on CIFAR-100 with a ResNet-18 backbone. Values for Adv-BNN, PNI and L2P are approximations extracted from~\citet{cvpr20learn2perturb} Fig. 5 and 6 (supplementary material).}
%     \centering
%     \vskip 0.15in
%     \begin{tabular}{lccc}
%         \toprule
%         Method      & Clean & FGSM  & PGD \\
%         \midrule
%         IAAT        & \hphantom{$\sim$} \textbf{63.9} & - & \hphantom{$\sim$} 18.5 \\
%         Adv-BNN     & $\sim$ 58.0 & $\sim$ 30.0 & $\sim$ 27.0 \\
%         PNI         & $\sim$ 61.0 & $\sim$ 27.0 & $\sim$ 22.0 \\
%         L2P         & $\sim$ 50.0 & $\sim$ 30.0 & $\sim$ 26.0 \\
%         \alg{}      & \hphantom{$\sim$} 52.1 & \hphantom{$\sim$} \textbf{36.1} & \hphantom{$\sim$} \textbf{31.9} \\
%         \bottomrule
%     \end{tabular}
%     \label{tab:sota_compare_fgsm_pgd_cifar100}
% \end{table}

\begin{table}[t]
    \caption{Comparison of state-of-the-art SNNs for FGSM and PGD attacks on CIFAR-10 and CIFAR-100 with a ResNet-18 backbone. Performance of Adv-BNN, PNI and L2P extracted from~\citet{cvpr20learn2perturb}.}
    \centering
    \vskip 0.15in
    \resizebox{1.\linewidth}{!}{
    \begin{tabular}{lcccccc}
        \toprule
        & \multicolumn{3}{c}{CIFAR-10} & \multicolumn{3}{c}{CIFAR-100} \\
        Method      & Clean & FGSM  & PGD & Clean & FGSM  & PGD \\
        \midrule
        Adv-BNN     & 82.2 & 60.0 & 53.6 & $\sim$ 58.0 & $\sim$ 30.0 & $\sim$ 27.0 \\
        PNI         & 87.2 & 58.1 & 49.4 & $\sim$ 61.0 & $\sim$ 27.0 & $\sim$ 22.0  \\
        L2P         & 85.3 & 62.4 & 56.1 & $\sim$ 50.0 & $\sim$ 30.0 & $\sim$ 26.0  \\
        SE-SNN      & 92.3 & 74.3 & - & - & - & - \\
        IAAT        & - & - & - & 63.9 & - & 18.5 \\
        \alg{}      & \textbf{93.2} & \textbf{77.6} & \textbf{71.4} & \textbf{70.1} & \textbf{51.5} & \textbf{42.7}  \\
        \bottomrule
    \end{tabular}
    }
    \label{tab:sota_compare_fgsm_pgd_cifar}
\end{table}

\begin{table}[t]
    \caption{Comparison of state-of-the-art SNNs for white box C\&W attack and black box n-Pixel attack on CIFAR-10 with a ResNet-18 backbone. Performance of competing methods extracted from~\citet{cvpr20learn2perturb}.}
    \centering
    \vskip 0.15in
    \resizebox{1.\linewidth}{!}{
    \begin{tabular}{llcccc}
        \toprule
         & Attack Strength & Adv-BNN & PNI & L2P & \alg{}  \\
         \midrule
         & Clean          & 82.2 & 87.2 & 85.3 & \textbf{93.2} \\
         \midrule
         \multirow{4}{*}{\rotatebox[origin=c]{90}{C\&W}} & $\kappa=0.1$ & 78.1 & 66.1 & 84.0 & \textbf{89.4} \\
         & $\kappa=1$     & 65.1 & 34.0 & 76.4 & \textbf{78.4} \\
         & $\kappa=2$     & 49.1 & 16.0 & 66.5 & \textbf{71.9} \\
         & $\kappa=5$     & 16.0 & 0.08 & 34.8 & \textbf{55.0} \\
         \midrule
         \multirow{4}{*}{\rotatebox[origin=c]{90}{n-Pixel}} & 1 pixel & 68.6 & 50.9 & 64.5 & \textbf{90.8} \\
         & 2 pixels   & 64.6 & 39.0 & 60.1 & \textbf{85.5} \\
         & 3 pixels   & 59.7 & 35.4 & 53.9 & \textbf{81.2} \\
         & 5 pixels   & - & - & - & 64.3 \\
         \bottomrule
    \end{tabular}
    }
    \label{tab:sota_compare_cw_1px}
\end{table}

% \begin{table}[t]
%     \caption{Comparison of state-of-the-art methods for One-Pixel Attack on CIFAR-10 with a ResNet-18 backbone. Performance of competing methods extracted from the supplementary material of~\citet{cvpr20learn2perturb}. Additionally, we report the performance of a 5-pixel attack.}
%     \centering
%     \vskip 0.15in
%     \resizebox{1.\linewidth}{!}{
%     \begin{tabular}{lcccc}
%         \toprule
%          Attack Strength & Adv-BNN & PNI & L2P & \alg{}  \\
%          \midrule
%          Clean      & 82.15 & \textbf{87.21} & 85.30 & 85.40 \\
%          1 pixel    & 68.60 & 50.90 & 64.45 & \textbf{81.98} \\
%          2 pixels   & 64.55 & 39.00 & 60.05 & \textbf{81.32} \\
%          3 pixels   & 59.70 & 35.40 & 53.90 & \textbf{77.72} \\
%          5 pixels   & - & - & - & 63.69 \\
%          \bottomrule
%     \end{tabular}
%     }
%     \label{tab:sota_compare_1_pixel}
% \end{table}

\begin{table}[t]
    \caption{Comparison of \alg{} to recent state-of-the-art, both stochastic and non-stochastic, on CIFAR-10. All competitors evaluate their models on the untargeted PGD attack, with attack strength $\epsilon=8/255$, and number of iterations $k \in \{7, 10, 20\}$. Some results are extracted from \citet{iccv19pni}. AT: Use of adversarial training.}
    \centering
    \vskip 0.15in
    \resizebox{1.\linewidth}{!}{
    \begin{tabular}{lllcc}
        \toprule
         Defense & Architecture & AT & Clean & PGD \\
         \midrule
         RSE \cite{eccv18rse} & ResNext                 & \xmark & 87.5 & 40.0 \\
         DP \cite{sp19dp} & 28-10 Wide ResNet           & \xmark & 87.0 & 25.0 \\
         TRADES \cite{icml19trades} & ResNet-18         & \checkmark & 84.9 & 56.6 \\
         PCL \cite{iccv19pcl} & ResNet-110              & \checkmark & 91.9 & 46.7 \\
         PNI \cite{iccv19pni} & ResNet-20 (4x)          & \checkmark & 87.7 & 49.1 \\
         Adv-BNN \cite{iclr19advbnn} & VGG-16           & \checkmark & 77.2 & 54.6 \\
         L2P \cite{cvpr20learn2perturb} & ResNet-18     & \checkmark & 85.3 & 56.3 \\
         MART \cite{iclr20mart} & ResNet-18             & \checkmark & 83.0 & 55.5 \\
         BPFC \cite{cvpr20bpfc} & ResNet-18             & \xmark & 82.4 & 41.7 \\
         RLFLAT \cite{iclr20rlflat} & 32-10 Wide ResNet & \checkmark & 82.7 & 58.7 \\
         MI \cite{iclr20mi} & ResNet-50                 & \xmark & 84.2 & 64.5 \\
         SADS \cite{cvpr20sads} & 28-10 Wide ResNet     & \checkmark & 82.0 & 45.6 \\
         \midrule
         \alg{} & ResNet-18                             & \xmark & \textbf{93.2} & \textbf{71.4} \\
         \bottomrule
    \end{tabular}
    }
    \label{tab:sota_compare_other}
\end{table}

\subsubsection{White-box Attacks} 
We first compare our proposed \alg{} to the existing state-of-the-art methods in the white-box attack setting. From the results in Table~\ref{tab:sota_compare_fgsm_pgd_cifar}, we can see that our~\alg{} shows noticeable improvement of $\sim 15\%$ over the strongest competitor, L2P. Moreover, we find that our method does not sacrifice its performance on clean data to afford such strong robustness.

An important aspect of WCA that needs to be assessed is its potential to scale with the number of classes. For this reason we conduct experiments on CIFAR-100, comparing against our previously mentioned competitors, plus IAAT~\cite{cvpr19denoising}, all of which use a ResNet-18 backbone in their architectures. From Table~\ref{tab:sota_compare_fgsm_pgd_cifar} we can see that the adversarial robustness of \alg{} outperforms the other methods.

We also present the evaluation of our method against the C\&W attack in Table~\ref{tab:sota_compare_cw_1px}. Here, the confidence level $\kappa$ indicates the attack strength. Our \alg{} achieves the best performance, with the accuracy degrading gracefully as the confidence increases.

\subsubsection{Black-box Attacks}
To further verify the robustness of our \alg{}, we conduct experiments on a black-box attack, the One-Pixel attack~\cite{ec19onepixel}. This attack is derivative-free and relies on evolutionary optimization, and its attack strength is controlled by the number of pixels it compromises. We follow~\citet{cvpr20learn2perturb} and consider pixel numbers in $\{1,2,3\}$. Additionally, we report results for a stronger 5-pixel attack. From Table~\ref{tab:sota_compare_cw_1px}, we can see that our method demonstrates the strongest robustness in all cases, showing $\sim 13\%$ to $\sim 22\%$ improvement over the best competitor Adv-BNN. Importantly, these results show that the robustness of our method does not rely on stochastic gradients.

\subsubsection{Stronger Attacks}
In addition, we evaluate \alg{} against two stronger attacks that are, in general, common among recent adversarial robustness literature, but are not mentioned in the stochastic defenses we outline as direct competitors. These are: (i) PGD$_{100}$; a stronger variant of PGD with 100 random restarts and (ii) the Square Attack~\cite{eccv20square}; a black-box attack that compromises the attacked image in small localized square-shaped updates. We present the results of our evaluation in Table \ref{tab:stronger_attacks}.

\begin{table}[t]
\caption{Evaluation of \alg{} with a ResNet-18 backbone on CIFAR-10, against the white-box PGD$_{100}$ and black-box Square Attack, for different values of attack strength $\epsilon$.}
\centering
\resizebox{1.0\columnwidth}{!}
{
  \begin{tabular}{clccccccccc}
    \toprule
    & $\epsilon/255$ & Clean & 1 & 2 & 4 & 8 & 16 & 32 & 64 & 128 \\
    \midrule
    \multirow{2}{*}{\rotatebox[origin=c]{90}{\scriptsize PGD$_{100}$}} & No Def.
    & 93.3 & 45.3 & 14.6 & 0 & 0 & 0 & 0 & 0 & 0 \\[1pt]
    & WCA & 93.2 & 73.2 & 72.2 & 72.1 & 71.2 & 69.7 & 56.4 & 28.2 & 10.5 \\[1pt]
    \midrule
    \multirow{2}{*}{\rotatebox[origin=c]{90}{\scriptsize Square}} & No Def. 
    & 93.3 & 32.9 & 31.7 & 12.4 & 6.0 & 1.2 & 0 & 0 & 0 \\[1pt]
    & WCA 
    & 93.2 & 51.7 & 51.7 & 50.4 & 49.0 & 48.8 & 44.3 & 36.9 & 28.6 \\[1pt]
    \bottomrule
  \end{tabular}
  \label{tab:stronger_attacks}
}
% \resizebox{1.0\columnwidth}{!}
% {
%   \begin{tabular}{lcccccccccc}
%     \toprule
%     $\epsilon/255$ & Clean & 1 & 2 & 4 & 8 & 16 & 32 & 64 & 128 \\
%     \midrule
%     PGD$_{100}$ & 93.2 & 73.2 & 72.2 & 72.1 & 71.2 & 69.7 & 56.4 & 28.2 & 10.5 \\
%     Square      & 93.2 & 51.7 & 51.7 & 50.4 & 49.0 & 48.8 & 44.3 & 36.9 & 28.6 \\
%     \bottomrule
%   \end{tabular}
%   \label{tab:stronger_attacks}
% }
\end{table}

\subsection{Comparison to State of the Art} 
Direct comparison to a wider range of competitors is difficult due to the variety of backbones and settings used. Nevertheless, Table~\ref{tab:sota_compare_other} provides comparison to recent state of the art stochastic and non-stochastic defenses. We can see that \alg{} achieves excellent performance including comparing to methods that use bigger backbones and make the stronger assumption of adversarial training.

% \begin{table*}[t]
%     \caption{Ablation study for FGSM and PGD attacks on CIFAR-10, CIFAR-100, SVHN and F-MNIST.}
%     \centering
%     \vskip 0.15in
%     \resizebox{1.\linewidth}{!}{
%     \begin{tabular}{lcccccccccccc}
%         \toprule
%          & \multicolumn{3}{c}{CIFAR-10} & \multicolumn{3}{c}{CIFAR-100} & \multicolumn{3}{c}{SVHN} & \multicolumn{3}{c}{F-MNIST} \\
%          \midrule
%          Model & Clean & FGSM & PGD & Clean & FGSM & PGD & Clean & FGSM & PGD & Clean & FGSM & PGD  \\
%          \midrule
%          No Defense                & 85.5 & 14.5 &  4.1 & 52.0 & 11.5 &  3.6 & 92.3 & 49.1 & 23.3 & 90.8 & 26.4 & 12.0 \\
%          \alg{} Isotropic       & 85.3 & 52.4 & 47.6 & 52.3 & 28.1 & 25.1 & 92.3 & 47.3 & 40.2 & 90.1 & 63.5 & 37.2 \\
%          \alg{} Anisotropic     & 85.4 & \textbf{72.4} & \textbf{71.7} & 52.3 & \textbf{36.1} & \textbf{31.9} & 92.1 & \textbf{72.6} & \textbf{66.7} & 90.1 & \textbf{65.2} & \textbf{48.5} \\
%          \bottomrule
%     \end{tabular}
%     }
%     \label{tab:comparison_isotropic_anisotropic}
% \end{table*}

\begin{table*}[t]
    \caption{Ablation study for FGSM and PGD attacks on CIFAR-10, CIFAR-100, SVHN and F-MNIST. For CIFAR-10, CIFAR-100 and SVHN we use a ResNet-18, and for F-MNIST a LeNet++ backbone.}
    \centering
    \vskip 0.15in
    \resizebox{1.\linewidth}{!}{
    \begin{tabular}{lcccccccccccc}
        \toprule
         & \multicolumn{3}{c}{CIFAR-10} & \multicolumn{3}{c}{CIFAR-100} & \multicolumn{3}{c}{SVHN} & \multicolumn{3}{c}{F-MNIST} \\
         \midrule
         Model & Clean & FGSM & PGD & Clean & FGSM & PGD & Clean & FGSM & PGD & Clean & FGSM & PGD  \\
         \midrule
         No Defense             & 93.3 & 14.9 &  3.9 & 72.2 & 12.3 &  1.2 & 93.4 & 55.6 & 23.5 & 90.8 & 26.4 & 12.0 \\
         \alg{} Isotropic       & 93.1 & 60.7 & 55.9 & 70.1 & 27.5 & 21.8 & 93.4 & 45.0 & 40.1 & 90.1 & 63.5 & 37.2 \\
         \alg{} Anisotropic     & 93.2 & 77.6 & 71.4 & 70.1 & 51.5 & 42.7 & 93.4 & 87.6 & 85.7 & 90.1 & 65.2 & 48.5 \\
         \bottomrule
    \end{tabular}
    }
    \label{tab:comparison_isotropic_anisotropic}
\end{table*}

\begin{table}[t]
    \caption{Control experiments on CIFAR-10 for further analysis. See Sec.~\ref{sec:further_analysis}. AT: Training purely with adversarial examples. CT+AT: Training with a mix of clean and adversarial examples.}
    \centering
    \vskip 0.15in
    \resizebox{1.\linewidth}{!}{
    \begin{tabular}{lccc}
        \toprule
        Experiment & Clean & FGSM & PGD \\
        \midrule
        No Defense & 93.3 & 14.9 & 3.9 \\
        \alg{} (Penalty regularizer) & 93.2 & 77.6 & 71.4 \\
        \alg{} (Constraint regularizer) & 92.2 & 62.9 & 53.2 \\
        \midrule
        E1: Test without EoT                & 93.2 & 82.9 & 75.1 \\
        E2: Average multiple noise samples  & 93.2 & 70.3 & 68.8 \\
        E3: Noise trained independently     & 93.1 & 45.0 & 41.6 \\
        \midrule
        \alg{}: AT      & 88.1 & 75.4 & 70.4 \\
        \alg{}: CT+AT   & 90.0 & 75.6 & 70.7 \\
        \bottomrule
    \end{tabular}
    }
    \label{tab:control_experiments}
\end{table}

\begin{table}[t]
    \caption{Comparison between the undefended ResNet-18 baseline and \alg{} with a ResNet-18 backbone for Imagenette (high-res, 10 categories) and mini-ImageNet (large-scale, 100 categories) under PGD attack.}
    \centering
    \vskip 0.15in
    \resizebox{1.\linewidth}{!}{
    \begin{tabular}{lcccccc}
        \toprule
        & \multicolumn{3}{c}{Imagenette} & \multicolumn{3}{c}{mini-ImageNet} \\
        Model      & Clean & FGSM & PGD & Clean & FGSM & PGD \\
        \midrule
        No Defense  & 75.5 &  8.4 & 0    & 51.9 &  5.0 & 0 \\
        \alg{}      & 74.2 & 59.3 & 48.7 & 51.3 & 41.6 & 30.4  \\
        \bottomrule
    \end{tabular}
    }
    \label{tab:imagenet_experiments}
\end{table}

\subsection{Further Analysis}
\label{sec:further_analysis}

\myparagraph{Ablation Study}
We perform an ablation study on four benchmarks, CIFAR-10, CIFAR-100, SVHN and F-MNIST, to investigate the contribution of anisotropic noise, as shown in Table~\ref{tab:comparison_isotropic_anisotropic}. For each benchmark, we evaluate a ``clean'' baseline architecture, consisting only of the backbone and the classification layer. We then evaluate a variant of \alg{} with isotropic, and one with anisotropic noise. We observe that our anisotropic noise provides consistent benefit to adversarial robustness.

Another important observation is that there is no trade-off between the robust and clean performance of our models; both the isotropic and anisotropic variants of \alg{} maintain the clean performance of the baseline defenseless model.

\begin{figure}[t]
    \centering
    \includegraphics[width=.49\columnwidth]{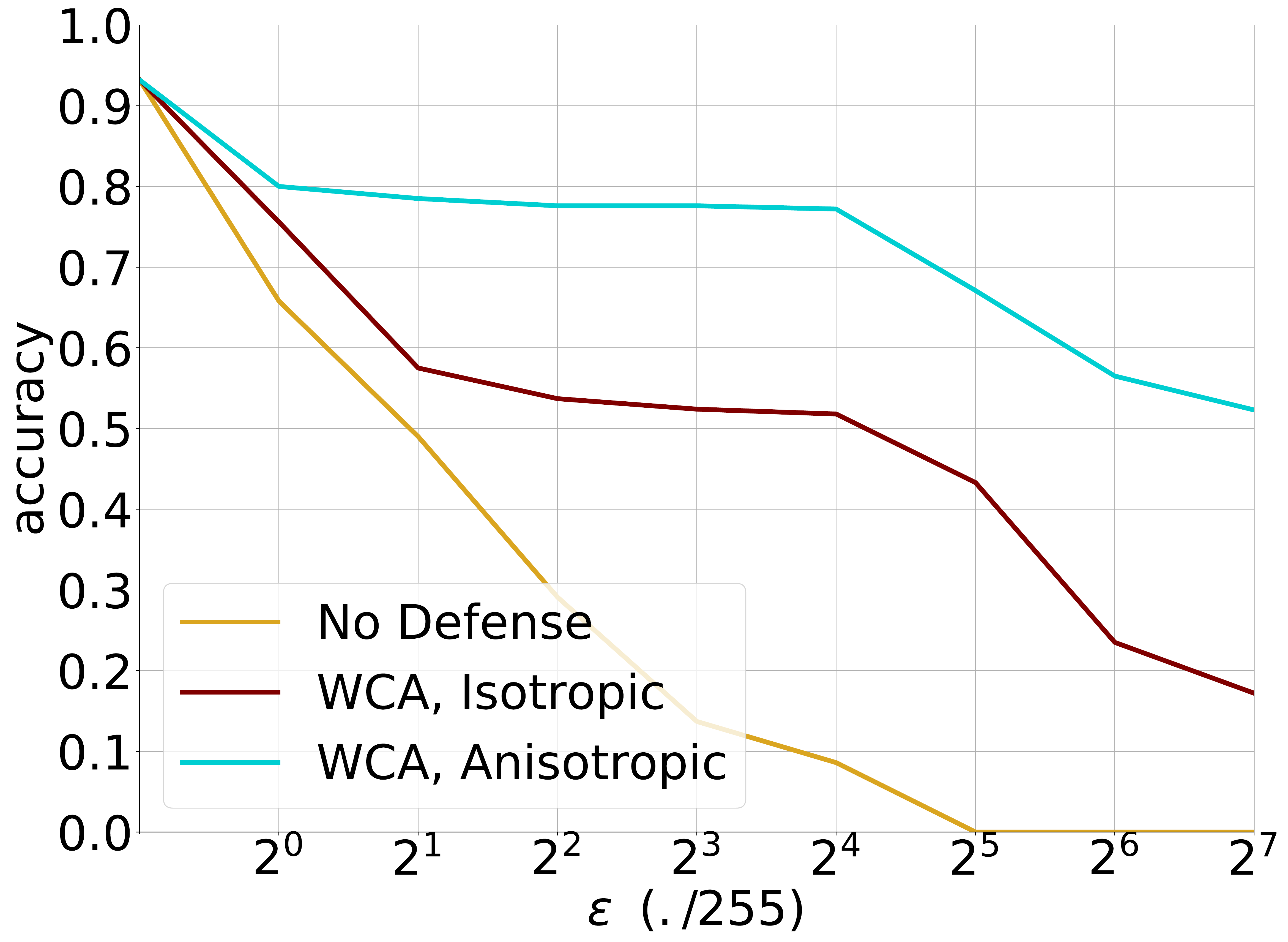}
    \includegraphics[width=.49\columnwidth]{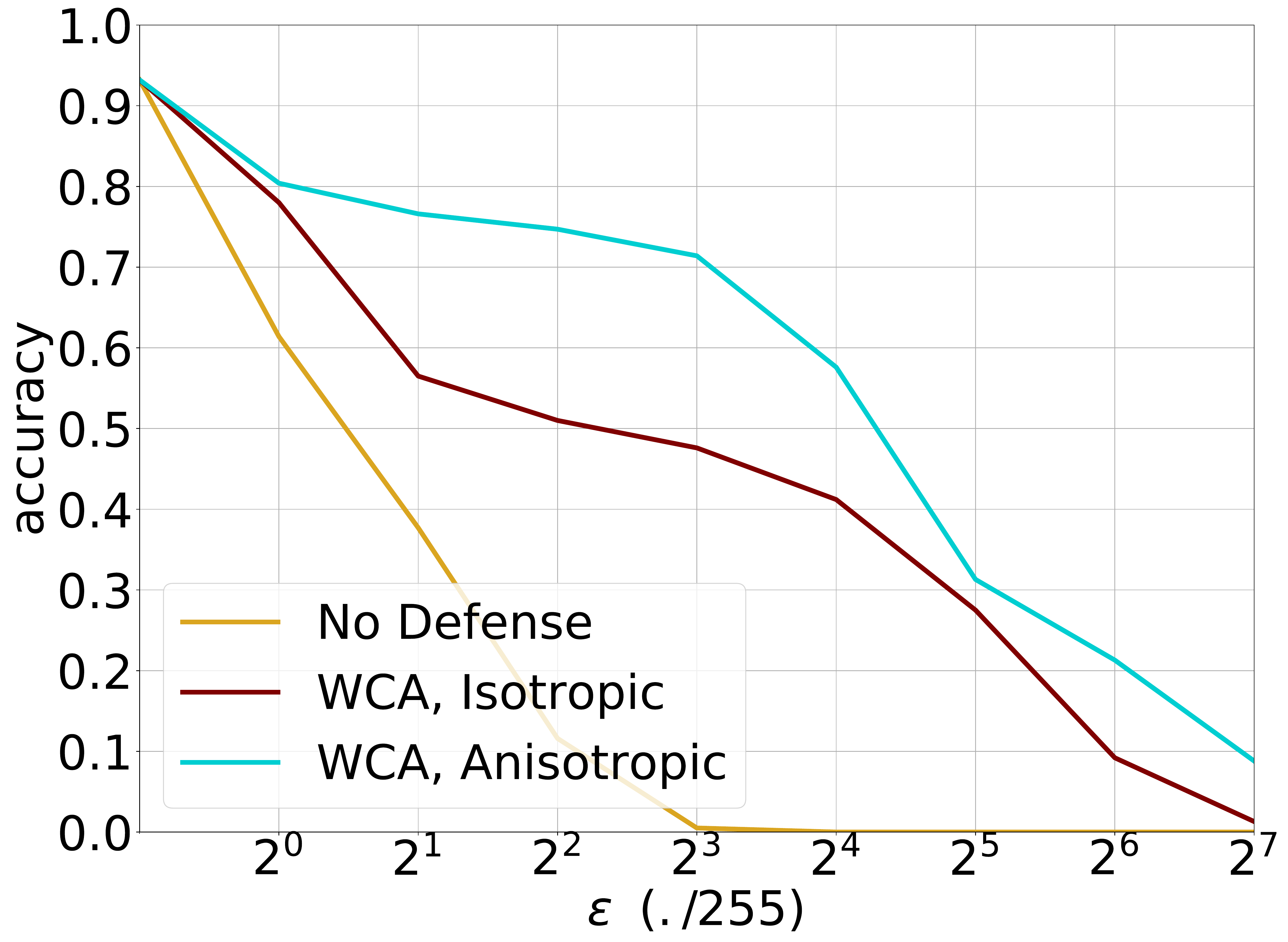}
    \caption{Evaluation of our model variants (see Table \ref{tab:comparison_isotropic_anisotropic}) for different attack strengths $\epsilon=2^n, \; n \in \{0...7\}$, specifically for the FGSM (left) and PGD (right) attacks on CIFAR-10. Best viewed in color.}
    \label{fig:ablation_epsilon}
\end{figure}

All the FGSM and PGD attacks in Table~\ref{tab:comparison_isotropic_anisotropic} use attack strength $\epsilon=8/255$. For completeness, we report the performance of all the variants above against FGSM and PGD with various attack strengths $\epsilon=2^n$, $n \in \{0...7\}$ on CIFAR-10 shown in Figure~\ref{fig:ablation_epsilon}. From these results, we can see the overall trend here is consistent with the observations in Table~\ref{tab:comparison_isotropic_anisotropic}. Also, we can see that the performance of our variants degrades more gracefully than the defenseless baseline.

\myparagraph{Large-scale, high-resolution} We are further interested to show that our \alg{} can handle high-resolution images and more challenging datasets. For that purpose we evaluate our method on two additional benchmarks: (i) Imagenette\footnote{\url{https://github.com/fastai/imagenette}}, a subset of ImageNet with 10 classes and full-resolution images, and (ii) mini-ImageNet~\cite{neurips16miniImagenet}, a large subset of ImageNet with 100 classes and 84x84 images, designed to be more challenging than CIFAR-100. The results presented in Table \ref{tab:imagenet_experiments} demonstrate that our method generalizes quite well to both high-resolution images as well as more challenging datasets.

\myparagraph{Norm-constrained architecture}
As explained in Section~\ref{sec:wca}, we control the magnitude of the weights in our architecture by means of $\ell^2$ regularization. Another option to achieve the same effect is to apply norm constraints to the classification vectors $\vec{w_i}$ and covariance matrix $\Sigma$. A detailed explanation of how we apply these norm constraints is given in the supplementary material. In Table~\ref{tab:control_experiments} we report results of a \alg{} variant with a norm-constrained regularizer. Constraint-based regularization still provides good robustness, but is weaker than the $\ell^2$ penalty-based variant.

\myparagraph{E1: Importance of EoT}
To show the impact of EoT, we also evaluate the test performance without it. Table~\ref{tab:control_experiments} shows that the test performance increases without using EoT. This makes sense as critiqued in ~\citet{pmlr18obfuscated}; one gradient sample is not enough to construct an effective attack.

\myparagraph{E2: Average multiple noise samples at test time}
Our model's forward pass performs the following: (i) Extract features from the penultimate layer of the backbone, (ii) inject additive noise, and (iii) compute the logits. By default we draw a single noise sample as suggested by our theory. In this experiment, we sample from the distribution multiple times and average the final logits. The more noise samples we average, the more we expect the additive noise to lose its regularization effect. The experimental results  in Table~\ref{tab:control_experiments} confirm that using more ($n=10$) samples degrades performance. 

\myparagraph{E3: Train noise and model independently}
In this experiment, we first train the model without injecting any noise. Then, keeping the model parameters frozen we train the noise independently. 
In Table~\ref{tab:control_experiments} we can see that this variant achieves an elementary level of robustness that is better than the defenseless baseline shown in Table~\ref{tab:comparison_isotropic_anisotropic}, however, not as strong as the isotropic baseline. As mentioned in Section~\ref{sec:wca}, a key insight of Theorem 1 is that the noise and weights should co-adapt. As expected, keeping the weight vectors $\vec w_i$ frozen, overall limits the ways the WCA term (see Eq.~\ref{eq:loss_wca}) can inflate, thus never realizing its full potential.

\myparagraph{Adversarial training}
Our proposed method only requires clean data for training. To show this, we adversarially train our anisotropic \alg{} in two settings: (i) purely with adversarial examples and (ii) with a mix of clean and adversarial examples. We train with a PGD attack with $\epsilon=8/255$ and $k=10$. From the results in Table~\ref{tab:control_experiments}, we can see that incorporating adversarial training harms our performance on clean data as expected~\cite{iclr15fgsm}; while providing no consistent benefit for adversarial defense.

\subsection{Inspection of Gradient Obfuscation}

\citet{pmlr18obfuscated} proposed a set of criteria to inspect whether a stochastic defense method relies on obfuscated gradients. Following~\citet{iccv19pni}, we summarize these criteria as a checklist. If any item in this checklist holds true, the stochastic defense is deemed unreliable. The following analysis verifies that our model's strong robustness is not caused by gradient obfuscation.

\myparagraph{Criterion 1:} One-step attacks perform better than iterative attacks.

\myparagraph{Refutation:} Knowing that PGD is an iterative variant of FGSM, we use our existing evaluation to refute this criterion. From the results in Tables \ref{tab:sota_compare_fgsm_pgd_cifar}, \ref{tab:comparison_isotropic_anisotropic} and \ref{tab:control_experiments}, we can see that our \alg{} performs consistently better against FGSM than against PGD.

\myparagraph{Criterion 2:} Black-box attacks perform better than white-box attacks.

\myparagraph{Refutation:} From Tables~\ref{tab:sota_compare_fgsm_pgd_cifar} and \ref{tab:sota_compare_cw_1px} we observe that FGSM and PGD outperform the 1-pixel attack. In Figure~\ref{fig:ablation_epsilon} we see the effect of increasing the attack strength on both white-box attacks, and they still outperform the stronger 2-, 3- and 5-pixel attacks.

\myparagraph{Criterion 3:} Unbounded attacks do not reach 100\% success.

\myparagraph{Refutation:} For fair comparison to previous work, FGSM and PGD in this paper are parameterized following~\citet{iccv19pni}. However, for this check we deliberately increase the attack strength of PGD to $\epsilon=255/255$ and number of iterations to $k=20$. We evaluate all of our models against this attack, and they achieve an accuracy of 0\%.

\myparagraph{Criterion 4:} Random sampling finds adversarial examples.

\myparagraph{Refutation:} To assess this, we hand-pick 100 CIFAR-10 test images that our model successfully classifies during standard testing (100\% accuracy), but misclassifies under FGSM with $\epsilon=8/255$ (0\% accuracy). For each of these test images, we randomly sample 1,000 perturbed images within the same $\epsilon$-ball, and replace the original image if any of the samples result in misclassification. We then evaluate our model on these 100 images to get a performance of 98\%.

\myparagraph{Criterion 5:} Increasing the distortion bound doesn't increase success.

\myparagraph{Refutation:} Figure~\ref{fig:ablation_epsilon} shows that increasing the distortion bound increases the attack's success.

\subsection{Empirical Evaluation of Theorem~\ref{thm:bound}}

\begin{figure}[t]
    \centering
    \includegraphics[width=.49\columnwidth]{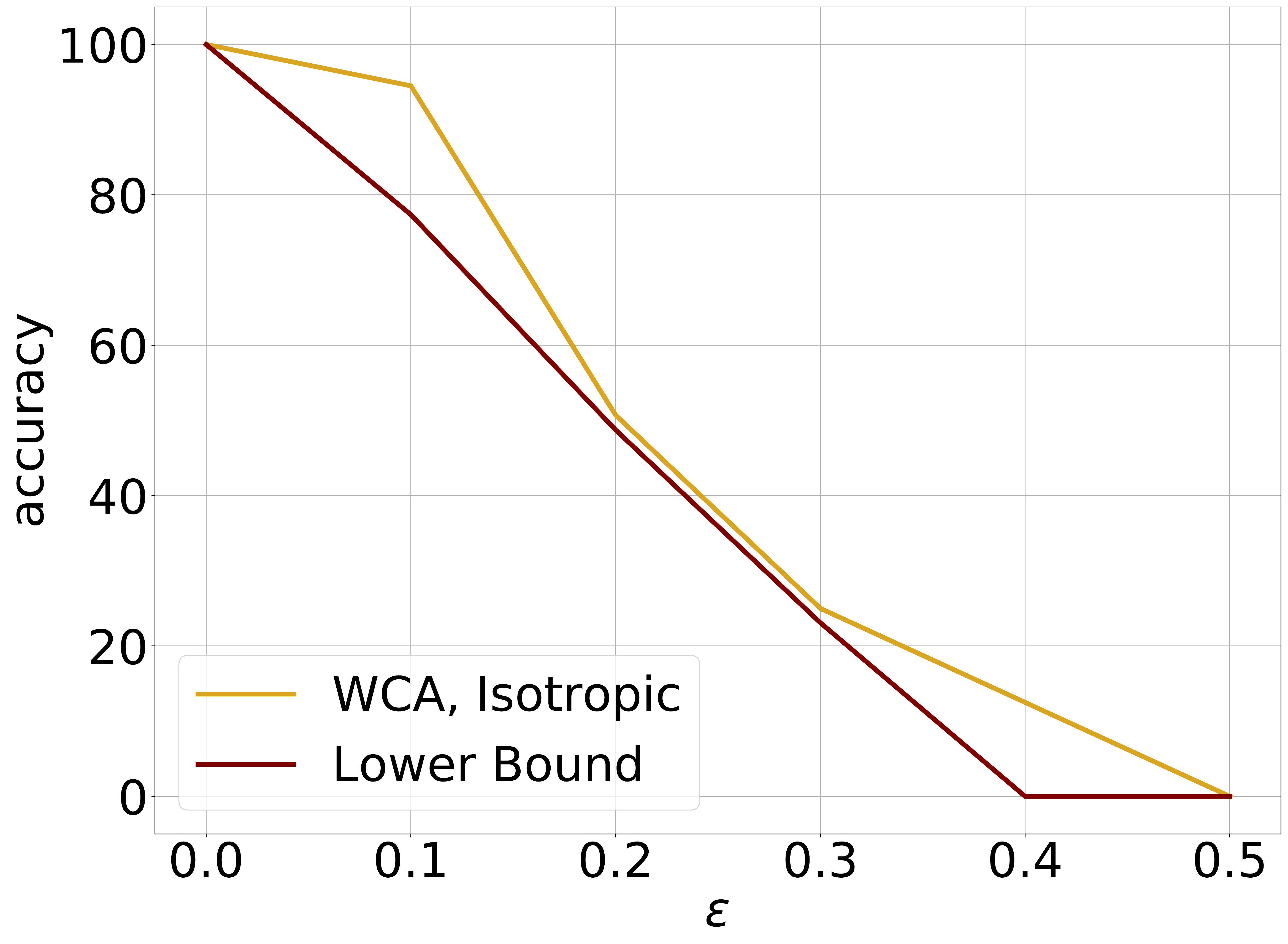}
    \includegraphics[width=.49\columnwidth]{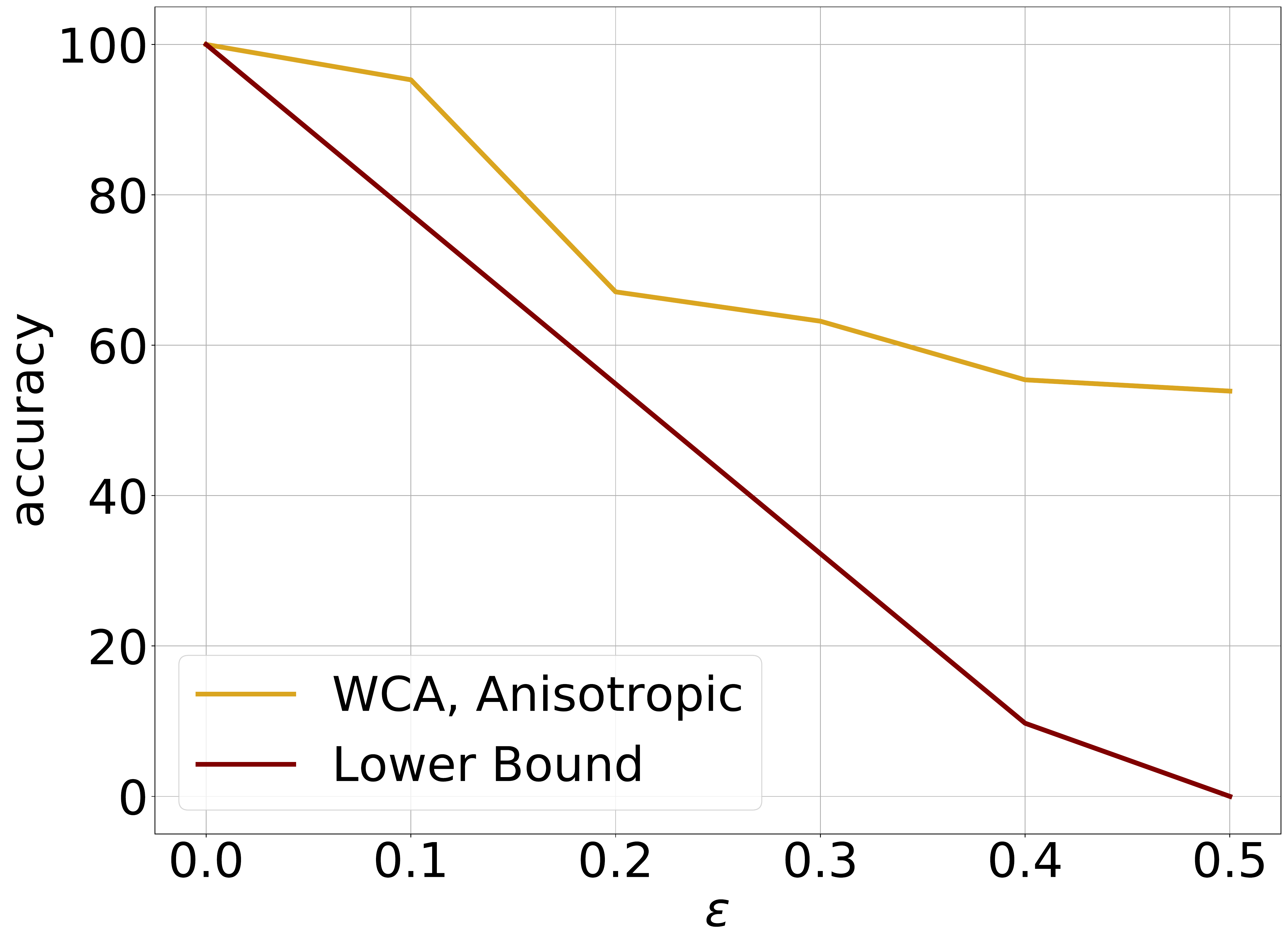}
    \caption{Evaluating our bound. Plots of the test set accuracy of SVMs trained on the zero and one digits found in MNIST. We report the performance of models trained with isotropic (left) and anisotropic (right) noise, and the worst-case performance according to Theorem~\ref{thm:bound}. The anisotropic model provides a more robust bound than the isotropic model as well as better empirical performance. Best viewed in color.}
    \label{fig:bound-plot}
\end{figure}

To evaluate the tightness of our bound presented in Theorem~\ref{thm:bound}, we train linear Support Vector Machines (SVM) on the zero and one digits found in the MNIST dataset. Using a linear model allows us to compute the numerator using the technique of \citet{gouk2020sspd},
\begin{equation*}
    \Delta_{\infty}^{\tilde{h}}(\vec x, \epsilon) = \epsilon \|\vec w\|_1,
\end{equation*}
where $\vec w$ is the weight vector of the SVM. We use principal components analysis to reduce the images to 32 dimensions, and apply learned isotropic and anisotropic noise to these reduced features before classification with the SVM. The covariance matrix and SVM weights are found by minimizing the hinge loss plus the WCA loss term using gradient descent. Results of attacking these models with PGD, and the lower bound on performance as computed by Theorem~\ref{thm:bound}, are given in Figure~\ref{fig:bound-plot}. From these plots we can see: (i) the bound is not violated at any point, corroborating our analysis; (ii) as the strength of the adversarial attack is increased, the bound remains non-vacuous for reasonable (i.e., likely imperceptible) values of the attack strength; and (iii) the model with anisotropic noise is more robust than the model with isotropic noise. This last finding is particularly interesting because in the linear model regime PGD attacks are able to find globally optimal adversarial examples.

\begin{figure}[t]
    \centering
    \includegraphics[width=.49\columnwidth]{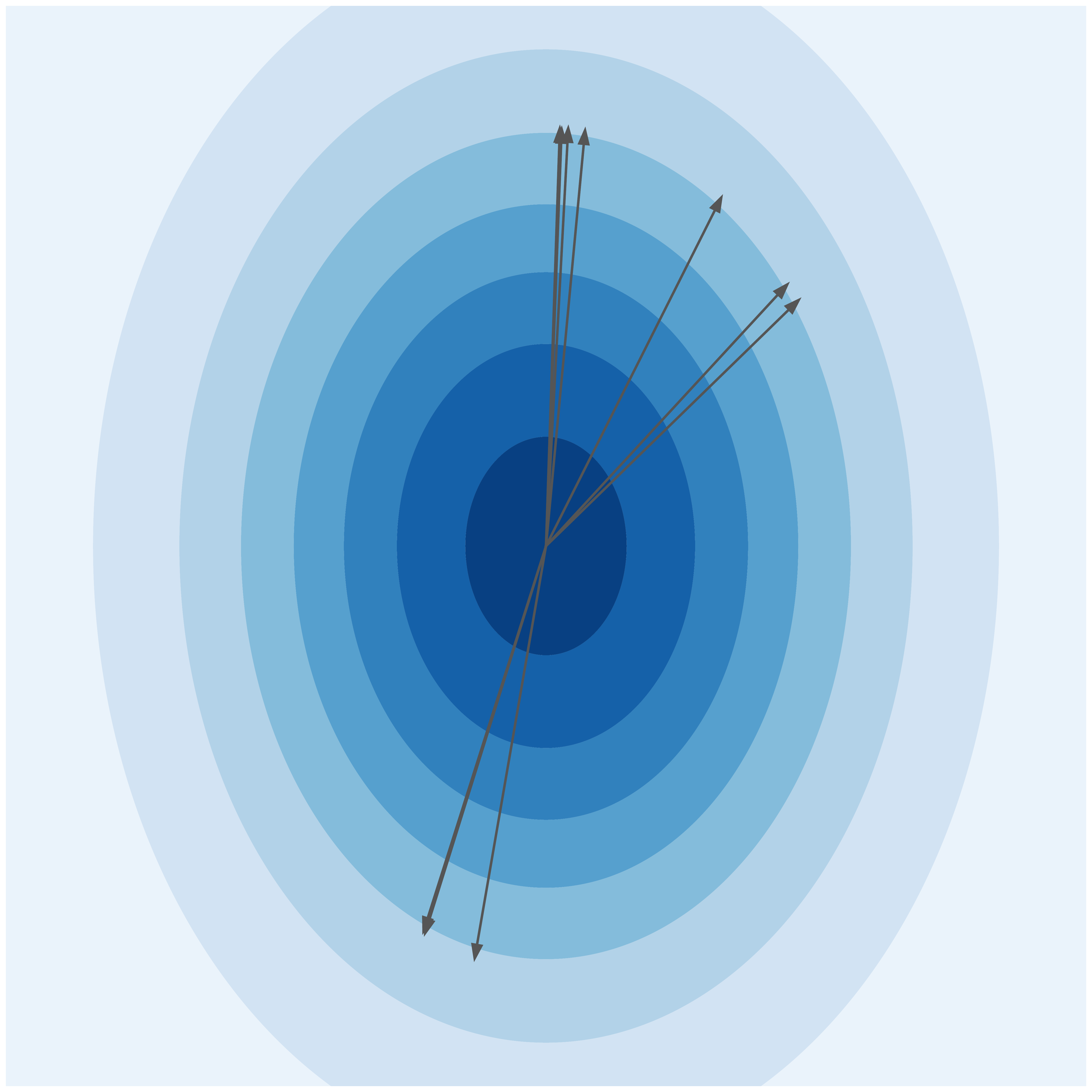}
    \includegraphics[width=.49\columnwidth]{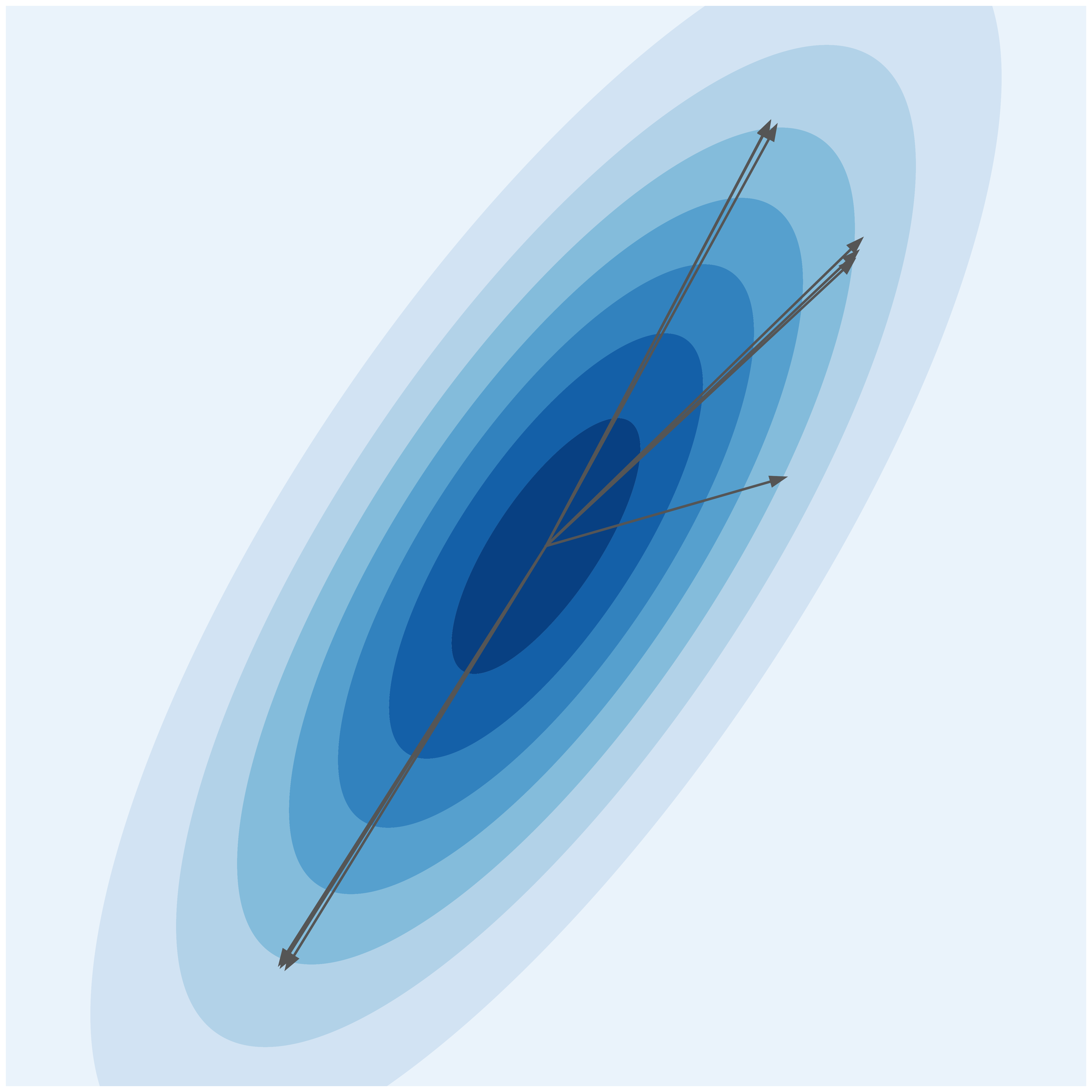}
    \caption{Visualisation of our models on F-MNIST with a 2D bottleneck. Contours and arrows indicate noise covariance $\Sigma$ and weights $ \vec w_i$. Left:  \alg{} with isotropic noise. Right: \alg{} with anisotropic noise. Evidently, our \alg{} with anisotropic noise allows covariance to be aligned with off-axis weights.}
    \label{fig:wca_contours}
\end{figure}

% \begin{figure}[t]
%     \centering
%     \includegraphics[width=.49\columnwidth]{Figures/grad_loss_vanilla.pdf}
%     \includegraphics[width=.49\columnwidth]{Figures/grad_loss_wca.pdf}
%     \caption{TBD}
%     \label{fig:grad_loss}
% \end{figure}

\subsection{Empirical Observations about WCA}

Figure~\ref{fig:wca_contours} shows the effect of our regularization methods with a bivariate Gaussian, by plotting the contours of the distribution against the weight vectors of the classification layer. These visualizations are obtained by training our \alg{} variants with a LeNet++ backbone on F-MNIST, with a 2-dimensional bottleneck and 2x2 covariance matrix.

We show the following: (i) First, in the left of Figure~\ref{fig:wca_contours}, we can see that the learned noise is axis-aligned since the injected noise is isotropic. Further, we can see that the weight vectors are near-axis-aligned, as WCA pushes them to align with the learned noise. 
(ii) Then, in the right Figure, due to the combination of anisotropic noise and WCA, our model has weight-aligned noise, and the weights are free to be non-axis-aligned. Overall, we observe better alignment between the learned weight vectors and the eigenvectors of the covariance matrix in our proposed anisotropic \alg{}.

\section{Conclusions}
\label{sec:conclusions}

In this paper we contribute the first stochastic model for adversarial defense that features fully-trained, anisotropic Gaussian noise, is hyperparameter free, and does not rely on adversarial training. We provide both theoretical support for the core ideas behind it, and experimental evidence of its excelling performance. We extensively evaluate \alg{} on a variety of white-box and black-box attacks, and further show that its high performance is not a result of stochastic (obfuscated) gradients. Thus, we consider the proposed model to push the boundary of adversarial robustness.

% \clearpage
% \newpage
\bibliography{main}
\bibliographystyle{icml2021}

% Appendix is not allowed, only in supplementary material. But let's keep it here for arXiv.
\clearpage
\appendix
\section{Proof of Theorem 1}

\begin{proof}
The definition of $h$ can be expanded to
\begin{equation*}
    h(\vec x) = \vec w^T f(\vec x) + \vec w^T \vec z + b, \;\; \vec z \sim \mathcal{N}(0, \Sigma),
\end{equation*}
and be reinterpreted as
\begin{equation*}
    h(\vec x) \sim \mathcal{N}(\vec w^T f(\vec x) + b, \vec w^T \Sigma \vec w).
\end{equation*}
Going further, we can see that the distribution of the margin function is
\begin{equation*}
    m_h(\vec x, y) \sim \mathcal{N}(y(\vec w^T f(\vec x) + b), \vec w^T \Sigma \vec w),
\end{equation*}
for which the probability of being less than zero is given by the cumulative distribution function for the normal distribution,
\begin{equation}
    \label{eq:theory_clean}
    P(m_h(\vec x, y) < 0) = \Phi \Bigg ( \frac{-y(\vec w^T f(\vec x) + b)}{\sqrt{\vec w^T \Sigma \vec w}} \Bigg ).
\end{equation}
From the increasing monotonicity of $\Phi$, we also have that
\begin{align*}
    \max_{\vec \delta : \|\vec \delta\|_p \leq \epsilon} & \Phi \Bigg ( \frac{-y(\vec w^T f(\vec x + \delta) + b)}{\sqrt{\vec w^T \Sigma \vec w}} \Bigg ) \\
    =& \Phi \Bigg ( \frac{\max_{\vec \delta : \|\vec \delta\|_p \leq \epsilon} -y(\vec w^T f(\vec x + \delta) + b)}{\sqrt{\vec w^T \Sigma \vec w}} \Bigg ).
\end{align*}
Suppose the adversarial perturbation, $\delta$, causes the output of the non-stochastic version of $h$ to change by a magnitude of $\Delta_p^{\Tilde{h}}(\vec x, \epsilon)$. There are a number of ways, such as local Lipschitz constants \cite{tsuzuku2018lipschitz,gouk2020sspd}, that can be used to bound the quantity for simple networks. Substituting $\Delta_p^{\Tilde{h}}$ into the previous equation yields
\begin{equation}
    \label{eq:theory_adv}
    \begin{split}
    \max_{\vec \delta : \|\vec \delta\|_p \leq \epsilon} & P(m_h(\vec x + \delta, y) \leq 0) \\
    \leq & \Phi \Bigg ( \frac{-y(\vec w^T f(\vec x) + b) + \Delta_p^{\Tilde{h}}(\vec x, \epsilon)}{\sqrt{\vec w^T \Sigma \vec w}} \Bigg ).
    \end{split}
\end{equation}
Finally, we know that the difference in probabilities of misclassification when the model is and is not under adversarial attack $\delta$, is given by
\begin{equation}
    \label{eq:theory_adv_gap_2}
    \begin{split}
    G_{p,\epsilon}^h(\vec x, y) = \max_{\vec \delta : \|\vec \delta\|_p \leq \epsilon} P(m_h(\vec x + \delta, y) \leq 0)\\ - P(m_h(\vec x, y) \leq 0).
    \end{split}
\end{equation}
Combining Equations~\ref{eq:theory_clean} and~\ref{eq:theory_adv} with Equation~\ref{eq:theory_adv_gap_2} results in
\begin{equation*}
    \begin{split}
    G(\vec x, y) \leq \Phi \Bigg ( \frac{-y(\vec w^T f(\vec x) + b) + \Delta_p^{\Tilde{h}}(\vec x, \epsilon)}{\sqrt{\vec w^T \Sigma \vec w}} \Bigg ) \\- \Phi \Bigg ( \frac{-y(\vec w^T f(\vec x) + b)}{\sqrt{\vec w^T \Sigma \vec w}} \Bigg ).
    \end{split}
\end{equation*}
Because the Lipschitz constant of $\Phi$ is $\frac{1}{\sqrt{2 \pi}}$, we can further bound $G$ by
\begin{equation*}
    \label{eq:theory_bound}
    G(\vec x, y) \leq \frac{\Delta_p^{\Tilde{h}}(\vec x, \epsilon)}{\sqrt{2 \pi \vec w^T \Sigma \vec w}}.
\end{equation*}
\end{proof}

\section{Hyperparameters of Experiments}

In Table~\ref{tab:hyperparams}, we provide the hyperparameter setup for all the experiments in our ablation study. Note that we use the same values for both the isotropic and anisotropic variants of our model within the same benchmark. We further clarify that we use a batch size of 128 across all experiments.
To choose these values, we split the training data into a training and a validation set and performed grid search. The grid consisted of negative powers of 10 \{$\mathrm{10^{-1}, 10^{-2}, 10^{-3}, 10^{-4}}$\} for both hyperparameters.

\begin{table}[t]
    \caption{Values for learning rate and weight decay for all experiments in our ablation study.}
    \centering
    \begin{tabular}{lcc}
        \toprule
        Benchmark & Learning rate & Weight decay \\
        \midrule
        CIFAR-10 & $10^{-2}$ & $10^{-4}$ \\
        CIFAR-100 & $10^{-2}$ & $10^{-4}$ \\
        SVHN & $10^{-2}$ & $10^{-4}$ \\
        FMNIST & $10^{-4}$ & $10^{-4}$ \\
        \bottomrule
    \end{tabular}
    \label{tab:hyperparams}
\end{table}

\section{Larger Architectures}

In the main body of the paper we explore how our method scales with the size of the backbone's architecture by experimenting with LeNet++ (small, 60 thousand parameters) and ResNet-18 (medium, 11 million parameters). In Table~\ref{tab:wrn-34-10} we also provide some experimental results on CIFAR-10 with the much larger Wide-ResNet-34-10 architecture (46 million parameters)

\begin{table}[t]
\caption{PGD test scores on CIFAR-10 using WRN-34-10, for different values of attack strength $\epsilon$.}
\centering
\resizebox{1.0\columnwidth}{!}
{
  \begin{tabular}{lcccccccccc}
    \toprule
    PGD($\epsilon/255$)  & Clean & 1 & 2 & 4 & 8 & 16 & 32 & 64 & 128 \\
    \midrule
    No Defense  & 0.97 & 0.63 & 0.60 & 0.26 & 0.12 & 0 & 0 & 0 & 0 \\
    WCA-Net     & 0.97 & 0.80 & 0.80 & 0.77 & 0.73 & 0.70 & 0.34 & 0.10 & 0 \\
    \bottomrule
  \end{tabular}
  \label{tab:wrn-34-10}
}
\end{table}

\section{Enforcing Norm Constraints}

In Section~\ref{sec:wca} we elaborate on how we use an $\ell^2$ penalty to prevent the magnitude of the classifier vectors $\vec{w}$ and covariance matrix $\Sigma$ from increasing uncontrollably.
Another approach for controlling the magnitude of the parameters, is enforcing norm constraints after each gradient descent update, using a projected subgradient method.
The projected subgradient method changes the standard update rule of the subgradient method from
\begin{equation*}
    \vec \theta^{(t+1)} \gets \vec \theta^{(t)} - \alpha \nabla_{\vec \theta} \mathcal{L}(\vec \theta^{(t)}),
\end{equation*}
to
\begin{align*}
    \vec u^{(t)} &\gets \vec \theta^{(t)} - \alpha \nabla_{\vec \theta} \mathcal{L}(\vec \theta^{(t)}) \\
    \vec \theta^{(t+1)} &\gets \underset{\vec v \in \Omega}{\text{arg}\min} \, \|\vec v - \vec u^{(t)}\|_2^2, 
    % \label{eq:projection-definition}
\end{align*}
where $\Omega$ is known as the feasible set. In our case there are three sets of parameters: the feature extractor weights, the linear classifier weights, and the covariance matrix. No projection needs to be applied to the extractor weights, as they are unconstrained. The linear classifier weights have an $\ell^2$ constraint on the vector associated with each class, so their feasible set it an $\ell^2$ ball---there is a known closed form projection onto the $\ell^2$ ball (see, e.g., \citet{gouk2021iclr}). The feasible set for the covariance matrix is the set of positive semi-definite matrices with bounded singular values. This constraint can be enforced by performing a singular value decomposition on the updated covariance matrix, clipping the values to the appropriate threshold, and reconstructing the new projected covariance matrix~\citep{lefkimmiatis2013hessian}. The final algorithm is given by
\begin{align}
    Y^{(t)} &\gets \vec \Sigma^{(t)} - \alpha \nabla_{\Sigma} \mathcal{L}(\vec \phi^{(t)}, \vec w^{(t)}, L^{(t)}) \nonumber \\
    \vec u^{(t)}_i &\gets \vec w_i - \alpha \nabla_{\vec w_i} \mathcal{L}(\vec \phi^{(t)}, \vec w^{(t)}, L^t) \nonumber \\
    \vec \phi^{(t+1)} &\gets \vec \phi^{(t)} - \alpha \nabla_{\vec \phi} \mathcal{L}(\vec \phi^{(t)}, \vec w^{(t)}, L^t) \nonumber \\
    \vec w_i^{(t+1)} &\gets \frac{1}{\text{max}(1, \frac{\|\vec u_i^{(t)}\|_2}{\gamma})} \vec u_i^{(t)} \nonumber \\
    U^{(t)} S^{(t)} V^{(t)} &\gets Y^{(t)T}Y^{(t)} \label{eq:svd-step} \\
    \Sigma^{(t)} &\gets U^{(t)} \tilde{S}^{(t)} V^{(t)} \nonumber \\
    L^{(t+1)T}L^{(t+1)} &\gets \Sigma^{(t)}, \label{eq:cholesky-step}
\end{align}
where (\ref{eq:svd-step}) is performing a singular value decomposition, $\tilde{S}$ represents the clipped version of $S$, and (\ref{eq:cholesky-step}) is computing the Cholesky decomposition.

\section{Source Code and Reproducibility}

The source code is openly available on GitHub: \url{https://github.com/peustr/WCA-net}.

\end{document}